\pdfoutput=1
\documentclass[runningheads]{llncs}
\usepackage[T1]{fontenc}
\usepackage{graphicx}
\usepackage{booktabs}
\usepackage[misc]{ifsym}
\usepackage{cite}
\usepackage{subcaption}

\usepackage{wrapfig}
\usepackage{lipsum}
\usepackage{multirow}
\usepackage{times}
\usepackage{enumitem}
\usepackage{indentfirst}
\usepackage{latexsym}
\usepackage[normalem]{ulem}
\useunder{\uline}{\ul}{}
\usepackage{amsmath}
\usepackage{amssymb}
\usepackage{latexsym}
\usepackage{algorithm,algorithmic}
\usepackage{babel}
\usepackage{graphicx}
\usepackage{xcolor,colortbl}
\usepackage{xurl} % allow line breaks in links
\usepackage[hidelinks, breaklinks,
            colorlinks=true,
            linkcolor=black,
            citecolor=black,
            urlcolor=black]{hyperref}
\usepackage{cleveref}

\usepackage[T1]{fontenc}

\newcommand{\method}{\texttt{f-CBM}}
\usepackage[utf8]{inputenc}
\usepackage{bbm}
\usepackage{microtype}

\usepackage{inconsolata}

\title{Towards Faithful\\ Multimodal Concept Bottleneck Models}

\begin{document}

% ECML ---------------------------------------------------
%N.B.: Author information (both in the \author{} and \authorrunning{} command) should only be present in the Camera-Ready Version of your paper. The version that you initially submit for review, ought to be double-blind. So, when initially submitting your paper, use:
%\author{Author information scrubbed for double-blind reviewing}
\author{Pierre Moreau\inst{1}, Emeline Pineau Ferrand\inst{1}, Yann Choho\inst{1,3}, Benjamin Wong\inst{1}, Annabelle Blangero\inst{1}, Milan Bhan\inst{1,2}}

% \author{Anonymous Submission}
% You may leave out the orcidID information, if you want to.
% Use \corr to indicate the corresponding author. Note the spacing around the \corr command. Only one author can be the corresponding author.

%N.B.: comment out the \authorrunning{} command for the double-blind version of your paper submitted for review. Later, if your paper is accepted, use the command for the Camera-Ready Version.
\authorrunning{P. Moreau et al.}
% First names are abbreviated in the running head.
% If there is one author, write 'A.L. Benjamin'.
% If there are two authors, write 'A.L. Benjamin and C.C. Broadus Jr.'
% If there are more than two authors, '[...] et al.' is used.

\institute{Ekimetrics, F-75009 Paris, France 
\and
Sorbonne Université, CNRS, LIP6, F-75005 Paris, France 
\and
Télécom Paris, F-91120 Palaiseau, France}

\maketitle              % typeset the header of the contribution
% ECML ---------------------------------------------------
% old footnote for emails --------------------------------
% --- Unmarked footnotes (appear at the bottom of the first page) ---
% Temporarily remove the footnote marker
%\let\thefootnote\relax
%\footnotetext{\texttt{\{pierre.moreau, milan.bhan, annabelle.blangero\} @ekimetrics.com}}
% Restore numbering if you will use footnotes later
%\let\thefootnote\arabic
% old footnote for emails --------------------------------

\begin{abstract}
% Concept Bottleneck Models (CBMs) are interpretable models that route predictions through a layer of human-interpretable concepts. To be truly interpretable, CBMs must faithfully detect concepts: concepts must be properly identified, and the model must rely only on their intended semantics, without encoding extraneous information that leaks into final predictions. Existing methods mostly address concept detection and concept leakage separately, and typically at the cost of predictive accuracy. This work introduces \method\ a new CBM methodology that jointly improves concept detection and reduce leakage, with the use of a Kolmogorov Arnold layer for better concept detection and a differentiable leakage loss for leakage addressing. Experiments are conducted in the multimodal setting with vision-language backbone, demonstrating that the proposed approach applies seamlessly to both image and text data, making it versatile across modalities.

Concept Bottleneck Models (CBMs) are interpretable models that route predictions through a layer of human-interpretable concepts. While widely studied in vision and, more recently, in NLP, CBMs remain largely unexplored in multimodal settings. For their explanations to be faithful, CBMs must satisfy two conditions: concepts must be properly detected, and concept representations must encode only their intended semantics, without smuggling extraneous task-relevant or inter-concept information into final predictions, a phenomenon known as leakage. Existing approaches treat concept detection and leakage mitigation as separate problems, and typically improve one at the expense of predictive accuracy. In this work, we introduce \method, a faithful multimodal CBM framework built on a vision-language backbone that jointly targets both aspects through two complementary strategies: a differentiable leakage loss to mitigate leakage, and a Kolmogorov-Arnold Network prediction head that provides sufficient expressiveness to improve concept detection. Experiments demonstrate that \method\ achieves the best trade-off between task accuracy, concept detection, and leakage reduction, while applying seamlessly to both image and text or text-only datasets, making it versatile across modalities.

\end{abstract}

% \vspace{0.5cm}
\section{Introduction} 
\label{intro}

The field of eXplainable Artificial Intelligence (XAI)~\cite{xai_2_0_survey} aims to make the behavior of black-box models more interpretable. A common distinction in XAI separates (1) \textit{post hoc} explanation methods, which provide interpretability after a model has been trained, from (2) models that are interpretable by design~\cite{madsen2024interpretability}. Concept Bottleneck Models (CBMs)~\cite{cbm_intro} belong to the latter category. CBMs first map input representations to a set of human-interpretable high-level attributes, called \emph{concepts}, which are then used by a linear layer to produce the final prediction. This architecture enhances interpretability by grounding both the intermediate representations and the decision process in understandable terms.

While CBMs have been widely studied in computer vision~\cite{label_free_cbm,cbm_incremental,concepf_embedding_models,concept_xai_survey} and, to a lesser extent, in NLP~\cite{cbm_plm,cb_llm,bhan2025towards}, they remain largely unexplored in multimodal settings~\cite{concept_xai_survey}. Regardless of the modality, a central requirement for CBMs is that they produce faithful concept-based explanations. Faithfulness, can be defined as "\textit{how accurately an explanation reflects the true reasoning process of the model}"~\cite{jacovi2020towards}, a definition widely adopted in the literature~\cite{XAI_nlp_survey} and which we follow throughout this work. Unfaithful explanations can have serious consequences in critical domains~\cite{rudin2019stop} such as healthcare~\cite{farah2023assessment}, as they may lead to misplaced trust in erroneous model decisions. For a CBM to produce faithful explanations, concepts must be accurately detected and only represent the intended concept within the Concept Bottleneck Layer (CBL). However, standard CBM architectures face two key limitations in this regard: (1) they do not systematically guarantee the accuracy of concept detection~\cite{bhan2025towards}, and (2) they are prone to leakage~\cite{addressing_leakage_cbm}. Leakage occurs when unintended predictive information is encoded in the CBL. This can take two forms: task leakage, where concept representations smuggle task-relevant signals beyond their intended semantics, and inter-concept leakage, where each concept encodes information about other concepts beyond their natural correlations. In both cases, the faithfulness of CBM explanations is compromised. In this paper, we propose faithful Multimodal Concept Bottleneck Model (\method), a novel approach for training faithful multimodal CBMs (mCBMs). Our main contributions are as follows: 

\begin{enumerate}
    \item We introduce a CLIP-based multimodal CBM architecture.
    \item We provide a preliminary analysis highlighting the interplay between concept detection accuracy, task leakage, and inter-concept leakage in mCBMs.
    \item We propose a fully differentiable measure of task leakage that can be incorporated into the training loss, enabling a drastic reduction in leakage.
    \item We use a Kolmogorov-Arnold Network as a concept-to-class layer, offering greater expressiveness while preserving interpretability and improving concept detection.
\end{enumerate}

\begin{figure}[H]
    \centering
    \includegraphics[width=0.6\textwidth]{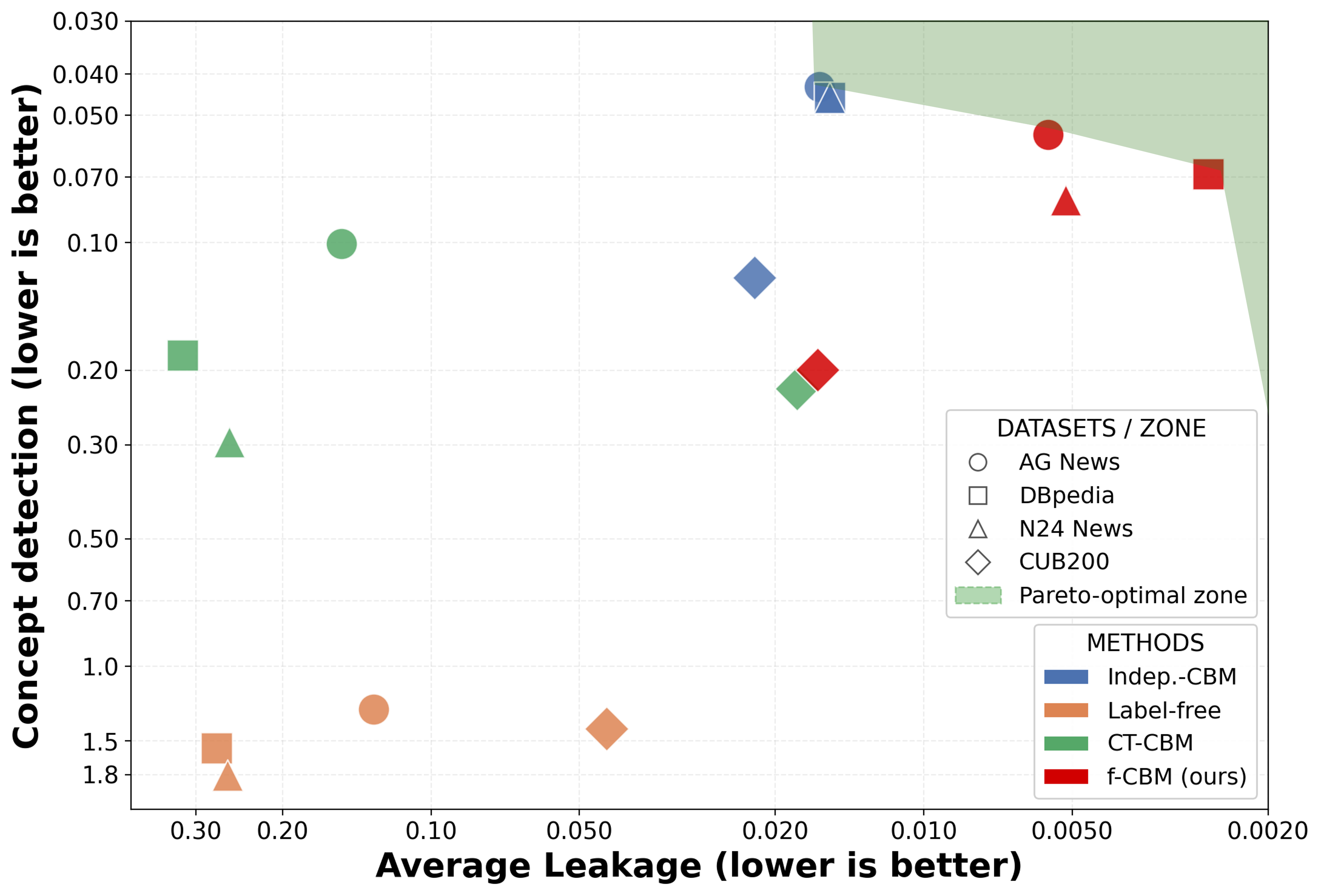}
    \caption{Pareto frontier: concept detection accuracy versus aggregate leakage. The x-axis represents the average of task-related and inter-concept leakage as introduced in Section~\ref{bk_rw}, and the y-axis represents RMSE concept detection performance.}
    \label{fig:pareto_cbm}
\end{figure}

Figure~\ref{fig:pareto_cbm} demonstrates that \method\ achieves the optimal trade-off between concept detection accuracy and leakage, positioning it on the Pareto frontier relative to competing approaches introduced in Section~\ref{bk_rw}. This paper is organized as follows:  Section~\ref{bk_rw} recalls some basic principles of concept-based interpretability, CBMs and related work. Section~\ref{sec:explanatory_anaysis} presents a preliminary analysis of faithfulness in a first multimodal CBM implementation, motivating the methodological choices introduced in the subsequent section. Section~\ref{sec:method} describes the proposed \method\ architecture in detail. Section~\ref{sec:xp} discusses our experiments, showing that \method\ systematically matches the performance of fine-tuned black-box multimodal models while accurately detecting concept activations and drastically reducing leakage. Finally, we illustrate the effect of leakage treatment on concept activations and show how the Kolmogorov-Arnold layer produces interpretable response curves, enabling concept-based explanations from \method.

\section{Background and Related Work}
\label{bk_rw}

This section first recalls some principles of Concept-based XAI and Concept Bottleneck Models in general. We then discuss related work on CBMs and multimodality and improving CBM faithfulness.

% This section reviews the development of Concept Bottleneck Models (CBMs), with a 
% particular focus on the phenomenon of concept leakage and the strategies proposed 
% to address it, before surveying existing extensions of CBMs to multimodal settings.

%Relevant articles to both multimodality and leakage, number of effective concepts \cite{srivastava2024vlgcbm}

%Fuzzy concepts \cite{NEURIPS2024_b5a41253}

\subsection{Background}

\paragraph{Concept-based XAI.}

The term \textit{concept} traces back to the Latin \textit{concipere}, "grasping something", and is understood in psychology as a mental entity enabling categorization and information integration~\cite{what_is_a_concept}. We ground our definition in Prototype Theory~\cite{rosch1975cognitive}, where concept are seen as a graded category membership (\textit{e.g.}, a toucan is more of a bird than a penguin). In the context of XAI, given an input text $x \in \mathcal{X}$ and a model $f$, concepts are defined as mapping from the input space to a value space~\cite{ruiz2024theoretical}. Formally, given a set of concepts $\mathcal{C}$ and a concept $c \in \mathcal{C}$, $c$ can be seen as the following operator: $c: \mathcal{X} \rightarrow \mathbb{R}$. Concept-based XAI then generates explanations either by assessing concept presence or by computing concept attributions, \textit{i.e.}, quantifying the importance of each concept to a given prediction~\cite{concept_xai_survey}.  

\paragraph{Concept Bottleneck Models.}

A CBM first maps the input to a vector of concept activations through a Concept Bottleneck Layer (CBL), then produces the final prediction via a linear layer. This sequential structure grounds the decision process in interpretable concepts. Three training regimes exist: \textit{joint} training optimizes concept detection and classification simultaneously, yielding the best task performance; \textit{independent} training supervises the CBL with concept labels and the classifier with ground-truth concepts, yielding the best concept accuracy at the cost of task performance; and \textit{sequential} training first freezes the CBL then trains the classifier on its outputs, combining the drawbacks of both. Beyond transparency, CBMs also enable \textit{concept intervention}, where users correct erroneous concept predictions at inference time to improve accuracy.

However, CBMs face several limitations: (1) the vast majority of CBM research focuses on vision~\cite{label_free_cbm,cbm_incremental,concepf_embedding_models,concept_xai_survey}, with only a few approaches addressing text~\cite{cbm_plm,cb_llm,bhan2025towards} and even fewer exploring multimodal settings, limiting their potential adoption across modalities, (2) most CBM architectures do not ensure the faithfulness of their concept-based explanations: concepts may be inaccurately detected in the CBL, or may encode unintended predictive information, a phenomenon known as leakage~\cite{mahinpei2021promises}. Moreover, leakage further undermines the value of concept intervention at inference time, as correcting concept activations can actually worsen predictions when the model relies on such unintended information~\cite{addressing_leakage_cbm}.

\subsection{Related Work}

\paragraph{Multimodality and Concept Bottleneck Models.}

While CBMs are well-established in computer vision and, more recently, in NLP, their extension to multimodal settings remains largely underexplored. The vision-language model CLIP~\cite{clip} was first embedded into CBMs for image classification, where concept importance is assessed by measuring the similarity between textual concept embeddings and image embeddings~\cite{yang2023language,label_free_cbm,kazmierczak2026enhancing}. These approaches exploit CLIP's contrastive training, which aligns large amounts of image-caption pairs in a shared embedding space, enabling concept detection in images without additional supervision. This framework has also been extended to other vision tasks such as retrieval~\cite{shi2025multimodal}. Another line of work~\cite{alukaev-etal-2023-cross} proposes aligning concept detection across text and images to improve visual concept detection robustness. However, this setup assumes that both modalities convey the same information and is ultimately used only for image classification. More recently, \cite{srivastava2024vlgcbm} improved concept detection by leveraging DINO-based grounding models~\cite{liu2024grounding} to provide visually grounded concept annotations, significantly enhancing annotation precision. While these works leverage multimodal models to improve unimodal vision CBMs, they remain limited to image classification, without addressing multimodal classification tasks.

% In \cite{alukaev-etal-2023-cross}, 
% textual and visual modalities are jointly used to fa to realistic datasets such as 
% CUB \cite{wah2011caltech}, promoting concept agreement across modalities; however, 
% the textual and visual components provide the same information and are handled by two separate models and the model is trained to align the concepts detected, rather than constructing a unified backbone.

\paragraph{Addressing CBM Faithfulness.} 

Several approaches have been proposed to improve CBM faithfulness, either by mitigating leakage or by improving concept detection reliability. A first strategy is to adopt an \textit{independent} training protocol, where the CBL is trained for concept detection alone, and the classification layer is subsequently trained on ground-truth concept values~\cite{cbm_intro}. This prevents the model from implicitly encoding unintended information in concept activations. We refer to this approach as \texttt{Independent-CBM} in the following. While it tends to minimize leakage and yield accurate concept detection, it comes at the cost of downstream task performance. An alternative approach~\cite{addressing_leakage_cbm} mitigates leakage by introducing a residual connection parallel to the CBM, which absorbs task-relevant information that concepts fail to capture. However, this side channel reduces interpretability, as part of the prediction bypasses the concept bottleneck. In NLP, \texttt{CT-CBM}~\cite{bhan2025towards} addresses faithfulness by carefully selecting a subset of concepts that can be reliably detected in the CBL. Leakage is handled through a similar residual connection, which is removed after training to restore interpretability. This results in improved concept detection with a drastically reduced concept set. More recently, \cite{parisini2025leakage} proposed a precise quantification of leakage in vision CBMs through the mutual information gain between task labels and concept representations, as well as between concept representations themselves. The authors define leakage in two forms: task leakage, where concept representations smuggle task-relevant signals beyond their intended semantics, and inter-concept leakage, where each concept encodes information about other concepts beyond their natural correlations~\cite{addressing_leakage_cbm}. The authors identify several causes of leakage, including insufficient concept supervision, an incomplete concept set, and a misspecified prediction head.

In this work, we address these limitations by proposing a multimodal concept bottleneck framework in which a single unified model processes both textual and visual modalities to predict concept activations. To improve faithfulness without significantly degrading task performance, we introduce two complementary strategies: a leakage-aware training objective derived from the quantification metrics proposed by \cite{parisini2025leakage}, and a more expressive yet interpretable prediction head. In the following section, we present a preliminary faithfulness analysis of a multimodal CBM implementation to motivate subsequent methodological choices.

\section{Preliminary Analysis: mCBM Faithfulness Factors Interplay}
\label{sec:explanatory_anaysis}

In this section, we conduct an exploratory analysis to better understand the interplay between what constitutes leakage in mCBMs trained following a classical joint training: concept detection performance, task leakage and inter concept leakage. We consider a corpus of multimodal samples $\mathcal{D} = \left\{ (x^v, x^t, y)\right\}$ 
where $x^v \in \mathcal{X}^v$ denotes the image, $x^t \in \mathcal{X}^t$ the associated 
text description, and $y \in \mathcal{Y}$ the label. $f^v : \mathcal{X}^v \rightarrow 
\mathbb{R}^d$ and $f^t : \mathcal{X}^t \rightarrow \mathbb{R}^d$ denote the vision and 
text encoders of a CLIP backbone \cite{clip}, respectively, with $d$ the 
dimension of the shared embedding space. For each sample, the two embeddings are 
concatenated into a single vector $z = [f^v(x^v) \| f^t(x^t)] \in \mathbb{R}^{2d}$, 
which serves as input to the CBL. We first introduce the dataset used in this preliminary study, then describe a straightforward multimodal CBM (mCBM) implementation for paired text and image inputs that achieves competitive task performance.

\subsection{A Baseline Joint mCBM Implementation}

\paragraph{Dataset} We use the N24News dataset~\cite{wang-etal-2022-n24news}, a multimodal news classification benchmark in which each sample pairs a news article (headline, abstract, caption, and body text) with an associated image, and the task is to predict the article category. Importantly, this dataset is constructed such that combining both modalities yields higher performance than using either one alone, making it a suitable testbed for multimodal CBMs. In our experiments, we use the abstract as the textual input, since the CLIP-base text encoder we use as a CBM backbone afterwards is limited to 77 tokens, favoring concise inputs.

\paragraph{Dataset Concept Annotation} The first step consists in enriching the dataset with concept labels to enable CBM training. Following~\cite{cb_llm}, we prompt \texttt{Claude 4.5 Sonnet} to generate a set of $k$ concepts $\mathcal{C} = \{c_1, c_2, \ldots, c_k\}$ relevant to discriminating between the target classes. Each modality is then independently annotated via cosine similarity against concept embeddings: image data is annotated using CLIP-Large~\cite{clip}, while textual data is annotated using a sentence transformer~\cite{reimers_sentence-bert_2019}. The resulting modality-specific concept scores are summed to produce a single unified concept vector per sample, yielding the enriched dataset $\mathcal{D}_c = \left\{ (x^v, x^t, \mathbf{c}, y)\right\}$ where $\mathbf{c}$ is the $k$-dimensional concept vector obtained by summing modality-specific concept scores. This dataset format is adopted throughout the remainder of the manuscript.

\paragraph{mCBM Training}We train the mCBM jointly using the following objective: $\mathcal{L} = \mathcal{L}_{\text{cls}}  + \lambda \, \mathcal{L}_{\text{C}}$ where $\mathcal{L}_{\text{cls}}$ is the cross-entropy loss, $\mathcal{L}_{\text{C}}$ is a mean squared error loss on concept prediction and $\lambda$ is an hyperparameter set to 1. Regarding optimization, the linear layers follow a cosine annealing learning rate schedule with an initial learning rate of either $10^{-1}$ or $10^{-2}$, while the CLIP backbone is fine-tuned with a fixed learning rate of $10^{-5}$. This configuration yields 98.1\% accuracy, closely matching the 98.5\% achieved by a fine-tuned CLIP model without a concept bottleneck. Having established that the mCBM achieves near-black-box performance, we now analyze its faithfulness on N24News.

\subsection{Studied mCBM Faithfulness Metrics}

We study the correlation between concept detection performance (measured by RMSE at the CBL) and two types of leakage following the measures proposed by~\cite{parisini2025leakage}, where concepts are denoted $c_i$ (ground-truth) and $\hat{c}_i$ (predicted by the mCBM). Concept-Task Leakage (CTL) measures whether predicted concepts encode unintended task-relevant information beyond what the true concepts capture:
\begin{equation}
\text{CTL}_i = \max\left(0, \frac{I(\hat{c}_i; y) - I(c_i; y)}{H(y)} \right),
\label{eq:CTL}
\end{equation}
where $I(\hat{c}_i; y)$ and $I(c_i; y)$ are mutual information between predicted/true concept and task label $y$, normalized by label entropy $H(y)$. Inter-Concept Leakage (ICL) quantifies unintended information sharing between concept pairs:
\begin{equation}
\text{ICL}_{ij} = \max\left(0, \frac{I(\hat{c}_i; \hat{c}_j) - I(c_i; c_j)}{\sqrt{H(\hat{c}_i) H(\hat{c}_j)}} \right),
\label{eq:ICL}
\end{equation}
measuring the excess mutual information between predicted concepts relative to their true correlations, normalized by marginal entropies. To analyze the relationship between concept detection performance and leakage, we discretize per-concept RMSE into three accuracy tiers: low, average, and high.

\begin{figure}[t]
    \centering
    \begin{subfigure}{0.48\textwidth}
        \centering
        \includegraphics[width=\textwidth]{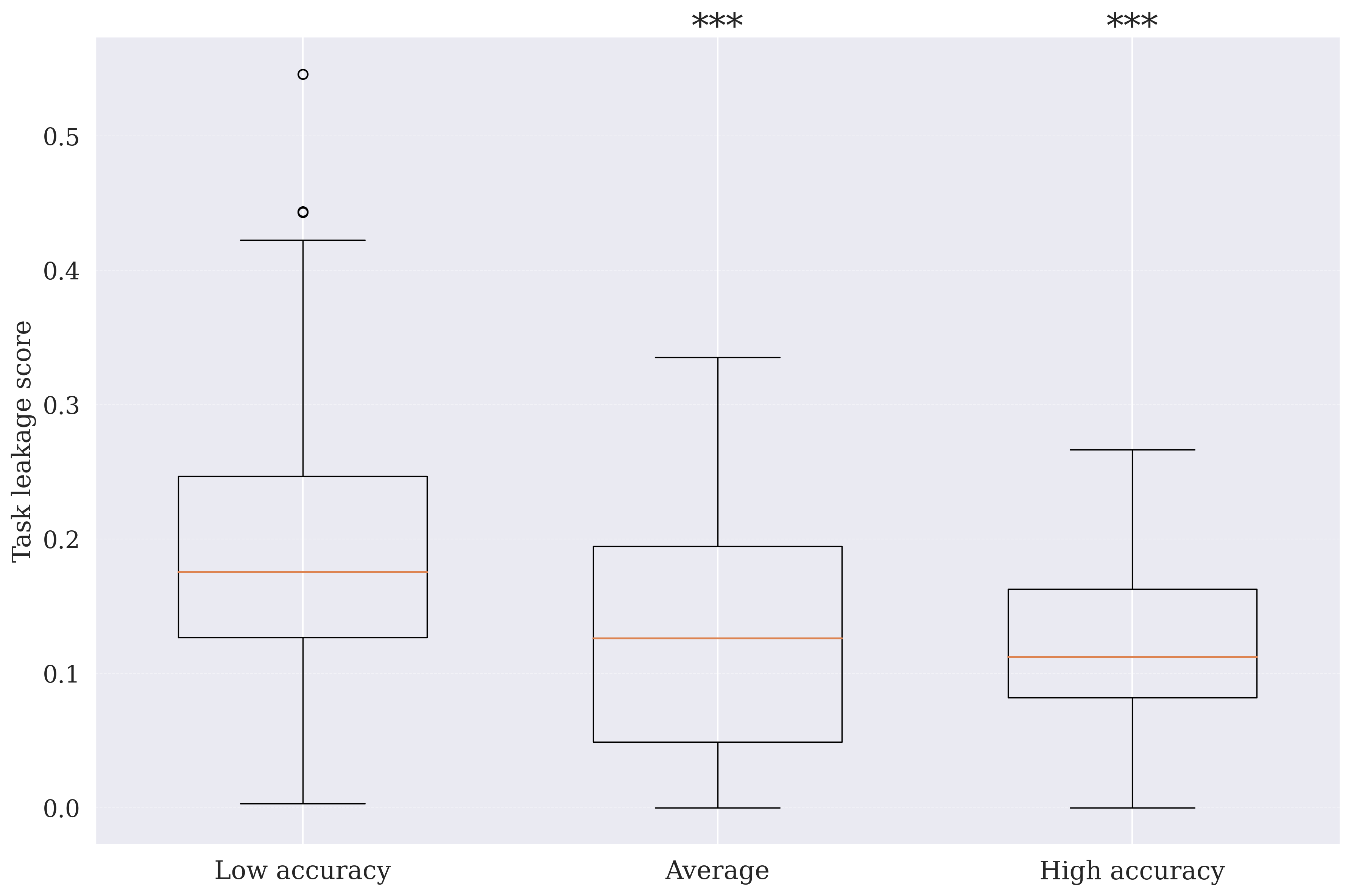}
        \caption{Task leakage vs. concept detection accuracy. With $p$ as the $p$-value of the one-tailed paired $t$-test, ***$p<1$\%. "Low accuracy" stands for the reference baseline.}
        \label{fig:task_leakage_boxplot_accuracy}
    \end{subfigure}
    \hfill
    \begin{subfigure}{0.48\textwidth}
        \centering
        \includegraphics[width=\textwidth]{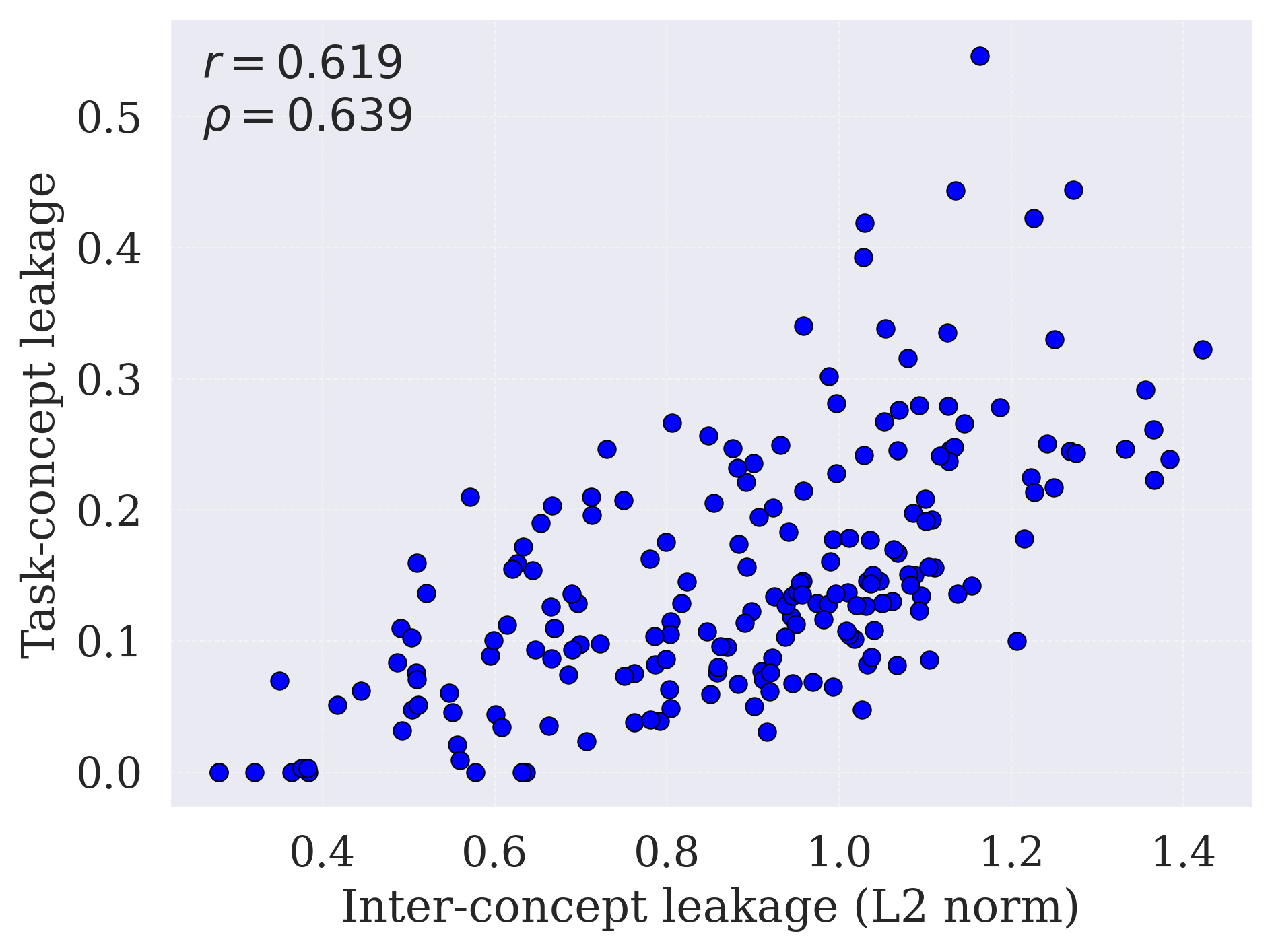}
        \caption{Correlation between inter-concept leakage \& task leakage}
        \label{fig:inter_vs_task}
    \end{subfigure}
    \caption{Leakage analysis in multimodal CBMs}
    \label{fig:leakage_combined}
\end{figure}

\subsection{Findings}

Figure~\ref{fig:task_leakage_boxplot_accuracy} shows that, within our jointly-trained mCBMs, concepts with moderate and high detection accuracy exhibit significantly lower task leakage (CTL) compared to poorly detected concepts, as confirmed by a paired-$t$-test. This results suggests a positive association between concept detection performance and reduced task-related leakage. Figure~\ref{fig:inter_vs_task} reveals a strong positive correlation between inter-concept leakage (ICL) and task-related leakage (CTL), confirmed by both Pearson and Spearman correlation coefficients ($r$, $\rho$). This suggests that task leakage and inter-concept leakage tend to co-occur: concepts that exhibit higher task leakage also tend to share more unintended information with other concepts.
 
Taken together, these observations point to a relationship between the constituents of faithfulness: concept detection accuracy, task-related leakage, and inter-concept leakage appear to be linked in jointly-trained mCBMs. Based on these findings, we hypothesize that a training strategy jointly targeting concept detection quality and task leakage reduction may also mitigate inter-concept leakage, leading to overall improved faithfulness. We leverage this hypothesis in the next section to develop faithful multimodal CBMs, and verify it empirically in Section~\ref{sec:xp}.

\section{Our method: \method\ }
\label{sec:method}

\begin{figure*}[t]{\centering}
\begin{center}
\includegraphics[width=0.99\textwidth]{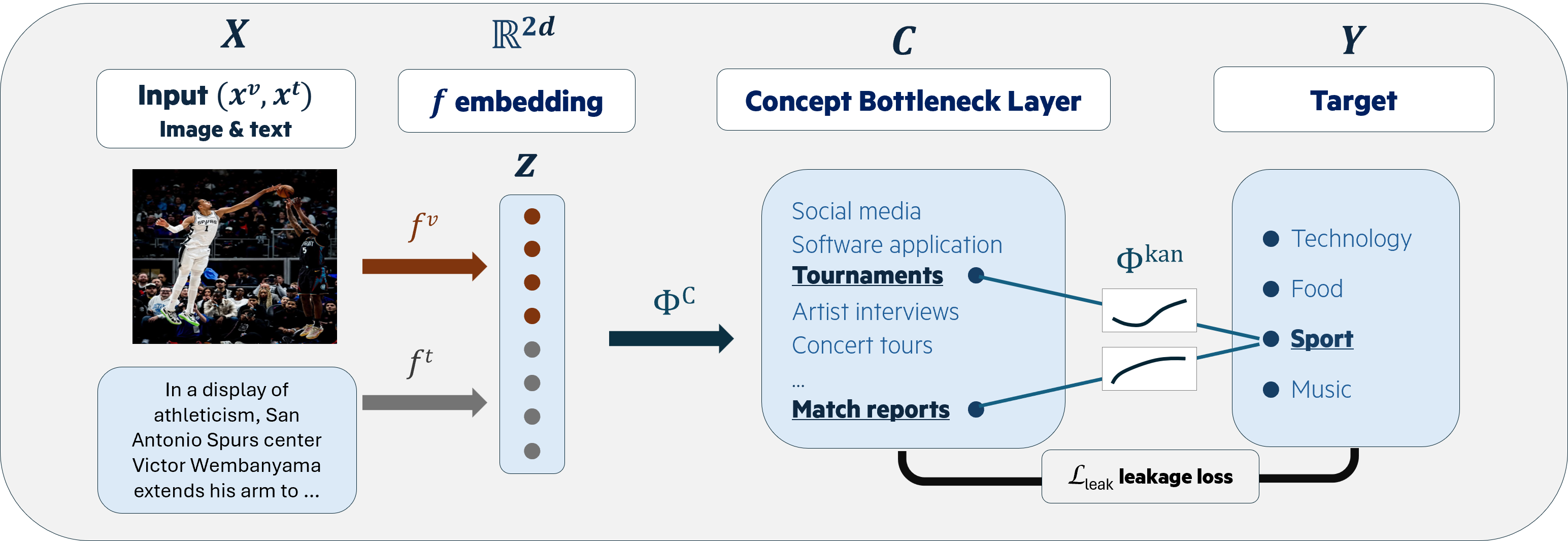}
\caption{Overview of \method, illustrated on an instance from the N24News dataset belonging to the Sport category.}
\label{fig:global_flow}
\end{center}
\end{figure*}

This section presents the architecture of the proposed framework, which addresses leakage through two complementary strategies: explicitly training the model to minimize leakage, and enhancing predictive flexibility at the final layer by replacing the conventional linear layer with a Kolmogorov–Arnold Network (KAN) layer, without sacrificing interpretability. Our \method\ architecture is summarized in Figure~\ref{fig:global_flow} and relies on the following components: \begin{enumerate}
    \item \textbf{Dataset Annotation.} As explained in Section~\ref{sec:explanatory_anaysis}, the dataset is annotated in terms of numerical concept values encompassing both modalities.
    \item \textbf{Leakage-Aware CBM Training.} The mCBM is trained with an additional loss that explicitly minimizes task-related leakage, ensuring concepts remain interpretable.
    % \item \textbf{Leakage-Aware CBM Training.} Given the annotated concept activations, the mCBM is trained by explicitly accounting for concept task leakage. Leakage is constrained throughout training by an additional loss, ensuring that predictions rely on interpretable concept activations rather than spurious information.
    \item \textbf{KAN Prediction Layer.} The conventional linear mapping from concept 
activations to final class predictions is replaced by a Kolmogorov-Arnold Network 
(KAN) layer \cite{liu2025kankolmogorovarnoldnetworks}, which offers greater predictive flexibility while preserving the interpretability of the concept bottleneck.
\end{enumerate}

The dataset annotation step having already been presented in Section~\ref{sec:explanatory_anaysis}, we directly present how we address leakage during the training.

\subsection{Leakage loss}

In order to train our model with an explicit objective to reduce leakage overall, we focus on minimizing concept task leakage, as reducing it should also reduce inter-concept leakage, as demonstrated in Section~\ref{sec:explanatory_anaysis}. To this end, we employ a differentiable estimator based on Kernel Density Estimation (KDE). %Following \cite{parisini2025leakage}, the information leakage is defined as:
% \begin{equation}
% \text{CTL} = \max\left(0, \frac{I(\hat{c}_i; y) - I(c_i; y)}{H(y)} \right),
% \end{equation} 
% where $I(\hat{c}_i; y)$ and $I(c_i; y)$ denote the mutual information between learned/true concepts and task labels, respectively, normalized by label entropy $H(y)$.
Unlike discrete binning approaches, which destroy gradient information, we estimate mutual information in Eq. \ref{eq:CTL} using Kernel Density Estimators~\cite{PhysRevE.52.2318,pmlr-v4-suzuki08a}:
\begin{equation}
    \hat{I}(x;y) = N^{-1} \sum_i \log\left[\frac{\hat{p}(x_i|y_i)}{\hat{p}(x_i)} \right]
\end{equation}
where $N$ is the number of samples and $\hat{p}(x_i) = (N-1)^{-1} \sum_{j \neq i} K_\sigma(x_i - x_j)$ is the probability density for the continuous variable $x$ using Gaussian kernels $K_\sigma(u) = (2\pi\sigma^2)^{-1/2} \exp(-u^2/2\sigma^2)$ with $\sigma$ calculated via Scott's rule $\sigma = 1.06 \cdot \text{std}(x) \cdot N^{-1/5}$~\cite{scott2015multivariate}. For the conditional density $\hat{p}(x_i | y_i)$, we first estimate $\hat{p}(x_i | y=Y)$ for each class $Y$ separately:
\begin{equation}
\hat{p}(x_i | y=Y) = (N_Y - \delta_{i,Y})^{-1} \sum_{j \in \{y_j = Y\},\, j \neq i } K_\sigma(x_i - x_j),
\end{equation}
where $N_Y$ is the number of samples in class $Y$ and $\delta_{i,Y}$ indicates whether sample $i$ belongs to class $Y$ to take into account self-exclusion. We then extract the class-specific density for each sample as $\hat{p}(x_i | y_i) = \hat{p}(x_i | y=y_i)$, selecting the density estimate corresponding to sample $i$. We employ a squared differentiable CTL loss 
\begin{equation}
    \mathcal{L}_\text{leak} = \left[ \frac{\hat{I}(\hat{c}_i; y) - \hat{I}(c_i; y)}{H(y)} \right]^2
    \label{eq:CTL_loss}
\end{equation}
rather than clamping at zero, which provides gradient signal in both directions and encourages learned concepts to preserve all information present in true concepts while penalizing additional task leakage.

\subsection{KAN Layer}

Instead of a final linear layer linking concept representations to final task, we implement a Kolmogorov-Arnold Network (KAN) layer~\cite{liu2025kankolmogorovarnoldnetworks} based on basis function parameterization. This substitution aims to increase the expressiveness of the final prediction head, allowing better concept detection which should also reduce leakage as observed in Section \ref{sec:explanatory_anaysis}. Unlike traditional MLPs using fixed activation functions on nodes, KANs place learnable univariate functions on edges, inspired by the Kolmogorov-Arnold representation theorem~\cite{kolmogorov1957representations}. To maintain interpretability, we employ a single KAN layer:
\begin{equation}
\Phi^{\text{kan}}_o(x) = s_o \times \sum_{i=1}^{N} \phi_{i,o}(x), \qquad \phi_{i,o}(x) = \sum_{m=1}^{M} c_{i,o,m} \cdot B_m(x)
\label{eq:response_curve}
\end{equation}
where $s_o$ are learnable scaling factors, $N$ is the number of input features, $M$ is the number of basis functions, and each $\phi_{i,o}$ maps input feature $i$ to output $o$ as a linear combination of degree-1 triangular basis functions $B_m$ centered at each grid point.% with $M = G + k$ depending on grid size $G$ and spline order $k$.

% To maintain interpretability, we employ a single layer which calculates:
% \begin{equation}
% \Phi^{\text{kan}}_o(x) = s_o \times \sum_{i=1}^{N} \phi_{i,o}(x)
% \end{equation}
% where $s_o$ are learnable scaling factors and $N$ is the number of input features. Each univariate function $\phi_{i,o}(x)$ maping input feature $i$ to output feature $o$ is represented as a linear combination of basis functions:
% \begin{equation}
% \phi_{i,o}(x) = \sum_{m=1}^{M} c_{i,o,m} \cdot B_m(x)
% \label{eq:response_curve}
% \end{equation}
% where $M = G + k$ depends on grid size $G$ and spline order $k$, and $c_{i,o,m}$ are learnable coefficients. For efficiency purposes, the basis functions are degree-1 triangular functions centered at each grid point.

\subsection{Training}
For each sample, the CLIP vision and text embeddings are 
concatenated into $z = [f^v(x^v) \| f^t(x^t)] \in \mathbb{R}^{2d}$, which serves 
as input to the concept bottleneck layer $\Phi^{\text{C}}: \mathbb{R}^{2d} \rightarrow 
\mathbb{R}^{\lvert \text{C} \rvert}$, mapping the multimodal representation to concept 
activation scores. Final predictions are produced by a KAN layer 
$\Phi^{\text{kan}}: \mathbb{R}^{\lvert \text{C} \rvert} \rightarrow \mathcal{Y}$, 
replacing the conventional linear classifier to allow greater predictive 
flexibility while preserving interpretability. The full model is thus defined as 
$\Phi^{\text{kan}} \circ \Phi^{\text{C}} \circ [f^v \| f^t]$. Our CBM is trained jointly, optimizing concept detection and downstream classification simultaneously. The total training objective reads:
\begin{equation}
\label{eqn:loss_total}
  \mathcal{L} = \mathcal{L}_{\text{cls}} 
  + \tilde{\lambda} \, \mathcal{L}_{\text{C}} 
  + \tilde{\lambda}_{\text{leak}} \, \alpha \, \mathcal{L}_{\text{leak}}
\end{equation}
where $\mathcal{L}_{\text{cls}}$ is the cross-entropy classification loss, 
$\mathcal{L}_{\text{C}}$ is the concept loss (MSE), and 
$\mathcal{L}_{\text{leak}}$ is the concept task leakage loss. To ensure that each 
loss component remains on a comparable scale throughout training, each auxiliary loss is dynamically rescaled by its running mean relative to the classification loss: $\tilde{\lambda}_{\text{C}} = \frac{\lambda}
{\overline{\mathcal{L}}_{\text{C}}} \cdot \overline{\mathcal{L}}_{\text{cls}}$ and $\tilde{\lambda}_{\text{leak}} = \frac{\lambda_{\text{leak}}}
{\overline{\mathcal{L}}_{\text{leak}}} \cdot \overline{\mathcal{L}}_{\text{cls}}$
% \begin{equation}
% \tilde{\lambda}_{\text{C}} = \frac{\lambda}
% {\overline{\mathcal{L}}_{\text{C}}} \cdot \overline{\mathcal{L}}_{\text{cls}}, 
% \qquad
% \tilde{\lambda}_{\text{leak}} = \frac{\lambda_{\text{leak}}}
% {\overline{\mathcal{L}}_{\text{leak}}} \cdot \overline{\mathcal{L}}_{\text{cls}}
% \end{equation}
where $\overline{\mathcal{L}}_{\text{cls}}$, $\overline{\mathcal{L}}_{\text{C}}$, and $\overline{\mathcal{L}}_{\text{leak}}$ denote the respective running means over the training steps. The hyperparameters $\lambda$ and $\lambda_{\text{leak}}$ control the relative weight of the concept and leakage losses, respectively, and are both set to 1 by default. The factor $\alpha$ follows a cosine annealing schedule from 0 to 1 over training, progressively activating the leakage loss to avoid interfering with early concept learning. Linear layers use cosine annealing with an initial learning rate of $10^{-1}$ or $10^{-2}$, while the CLIP backbone is fine-tuned at a fixed rate of $10^{-5}$.

% The $\lambda$ hyperparameter controls the trade-off between concept and task loss, while $\lambda_{\text{leak}}$ controls the importance of the leakage penalty; both are set to 1 by default. The factor $\alpha$ progressively activates the leakage loss throughout training via a cosine annealing schedule, increasing from 0 at the first epoch to 1 at the final epoch. This progressive activation is motivated by the observation that leakage tends to develop throughout training as the model optimizes for classification performance, and that applying the leakage penalty too early interferes with concept learning and degrades concept prediction quality. Regarding optimization, the linear layers follow a cosine annealing learning rate 
% schedule with an initial learning rate of either $10^{-1}$ or $10^{-2}$, while the CLIP backbone is fine-tuned with a fixed learning rate of $10^{-5}$.

\section{Experimental Settings}
\label{sec:xp}

This section presents the experimental study conducted across two multimodal datasets, two text datasets, and two multimodal classifiers of different sizes, first comparing \method\ to several competitors. Next, we assess the impact of each component of our framework to reduce leakage and finally we illustrate of our CBM performs under intervention in comparison to other settings.

\subsection{Experimental Protocol}

\paragraph{Models, Datasets and Evaluation Metrics} We apply \method\ on two fine-tuned multimodal classifiers of different sizes: \texttt{CLIP-base} and  \texttt{CLIP-large}~\cite{clip}. In addition, \method\ is tested on the four following classification datasets. As explained in Section \ref{sec:explanatory_anaysis}, the \textbf{N24News} dataset consists of news articles, where we select the abstract and the associated images \cite{wang-etal-2022-n24news}. The annotation yields 195 concepts, and from the 24 available labels we select four target classes: Sports, Music, Food, and Technology. For the \textbf{CUB}-200-2011 bird classification dataset \cite{wah2011caltech}, we follow the concept processing procedure described in \cite{cbm_intro} to first curate the binary annotations, retaining a set of 112 bird attributes out of the 312 initially present. Since CUB does not natively include textual descriptions, we generate them synthetically from each instance's activated attributes, inspired from \cite{alukaev-etal-2023-cross}, then using a paraphraser \cite{chatgpt_paraphraser} to produce more fluent, natural language descriptions.% Specifically, up to 10 activated attributes are randomly selected and concatenated into a sentence and subsequently paraphrased using a paraphraser \cite{chatgpt_paraphraser} to produce more fluent, natural language descriptions. %An example of such generated description is: "The bird exhibits a multi-colored tail, solid breast, and a bill roughly the same length as its head, with brown upperparts and wings, black eyes, and a solid belly."
%Since off-the-shelf sentence transformers and CLIP lack the fine-grained sensitivity required to reliably detect CUB-specific bird attributes, we fine-tune CLIP-base contrastively on the binary dataset. 
To obtain finer-grained concept annotations, we contrastively train CLIP-base on the binary dataset with the objective to align image representations with their corresponding activated attributes.% — such as "this bird has a black throat". 
The resulting fine-tuned CLIP model, used solely for the annotation, is then employed to compute cosine similarities between concept texts and both image and text descriptions, yielding continuous concept annotations for the CUB dataset. To reduce computational complexity, we select 15 classes out of the 200 available in the CUB dataset. We also highlight the versatility of \method\ on two text-only datasets: \textbf{AGNews} and \textbf{DBpedia}, enriched with concept labels from \cite{bhan2025towards}. We follow a 4-metric evaluation protocol, with the first metric being the final classification task accuracy (\textbf{\%ACC}), second concept detection error (\textbf{c-RMSE}) and finally two metric to measure leakage, (\textbf{CTL}) from Eq. \ref{eq:CTL} and (\textbf{ICL}) from Eq. \ref{eq:ICL}.

\paragraph{\method\ and Competitors.}
\method\ is trained jointly using all available concepts and following the training protocol described in Section~\ref{sec:method}. We compare it against three baselines, all extended to the multimodal setting. \texttt{Indep.-CBM} follows the independent training regime with a standard linear prediction head and no leakage loss. \texttt{Label-free} extends~\cite{label_free_cbm,cb_llm} to multimodal inputs and is trained jointly with a linear layer using the \textit{cos cubed} concept loss proposed in the liminar paper instead of MSE. \texttt{CT-CBM} extends~\cite{bhan2025towards}, which selects a subset of concepts maximizing both detectability and task relevance, and uses a residual connection during training to absorb task-relevant information not captured by the bottleneck. \texttt{CT-CBM} is trained jointly with MSE loss and a linear layer.

\begin{table*}[t]
\caption{%
  \texttt{f-CBM} and competitors evaluation on four test sets and two backbone sizes.
  Best results per row are in \textbf{bold} and second best are \underline{underlined}.
  \texttt{Black-box} is excluded from ranking (marked {--}) as it produces no concept-level outputs.}
\label{tab:results}
\centering
\scriptsize
\setlength{\tabcolsep}{4pt}
\begin{tabular}{@{}cc|ccccc|ccccc@{}}
\toprule
\multicolumn{2}{c|}{\textbf{\begin{tabular}[c]{@{}c@{}}Model backbone\\(size)\end{tabular}}}
& \multicolumn{5}{c|}{\textbf{\begin{tabular}[c]{@{}c@{}}\texttt{CLIP-base}\\(110M)\end{tabular}}}
& \multicolumn{5}{c}{\textbf{\begin{tabular}[c]{@{}c@{}}\texttt{CLIP-large}\\(395M)\end{tabular}}} \\
\cmidrule(lr){3-7} \cmidrule(lr){8-12}
\textbf{Dataset} & \textbf{Metric}
& \texttt{\begin{tabular}[c]{@{}c@{}}Black\\-box\end{tabular}}
& \texttt{\begin{tabular}[c]{@{}c@{}}Indep.\\-CBM\end{tabular}}
& \texttt{\begin{tabular}[c]{@{}c@{}}Label\\-free\end{tabular}}
& \texttt{CT-CBM}
& \texttt{\begin{tabular}[c]{@{}c@{}}f-CBM\\(ours)\end{tabular}}
& \texttt{\begin{tabular}[c]{@{}c@{}}Black\\-box\end{tabular}}
& \texttt{\begin{tabular}[c]{@{}c@{}}Indep.\\-CBM\end{tabular}}
& \texttt{\begin{tabular}[c]{@{}c@{}}Label\\-free\end{tabular}}
& \texttt{CT-CBM}
& \texttt{\begin{tabular}[c]{@{}c@{}}f-CBM\\(ours)\end{tabular}} \\
\midrule
%% ---- AG News ----
\multirow{4}{*}{\textbf{AG News}}
& \%ACC $\uparrow$    & 90.6 & 86.6 & \textbf{91.0} & \textbf{91.0} & {\ul 90.6} & 91.5 & 86.2 & \textbf{91.7} & {\ul 91.2} & 90.7 \\
& c-RMSE $\downarrow$ & -- & \textbf{0.043} & 1.264 & 0.101 & {\ul 0.056} & -- & \textbf{0.042} & 1.426 & 0.146 & {\ul 0.068} \\
& CTL $\downarrow$    & -- & {\ul 0.028} & 0.212 & 0.244 & \textbf{0.005} & -- & 0.922 & {\ul 0.207} & 0.262 & \textbf{0.004} \\
& ICL $\downarrow$    & -- & \textbf{0.005} & 0.050 & 0.059 & {\ul 0.006} & -- & {\ul 0.004} & 0.047 & 0.079 & \textbf{0.002} \\
\midrule
%% ---- DBpedia ----
\multirow{4}{*}{\textbf{DBpedia}}
& \%ACC $\uparrow$    & 100.0 & 97.3 & 99.2 & \textbf{99.3} & {\ul 99.2} & 99.3 & 98.1 & \textbf{99.3} & {\ul 99.0} & 98.9 \\
& c-RMSE $\downarrow$ & -- & \textbf{0.045} & 1.560 & 0.184 & {\ul 0.069} & -- & \textbf{0.035} & 1.773 & {\ul 0.040} & 0.041 \\
& CTL $\downarrow$    & -- & {\ul 0.027} & 0.383 & 0.448 & \textbf{0.003} & -- & {\ul 0.042} & 0.317 & 0.089 & \textbf{0.007} \\
& ICL $\downarrow$    & -- & {\ul 0.004} & 0.162 & 0.191 & \textbf{0.002} & -- & {\ul 0.009} & 0.116 & 0.021 & \textbf{0.003} \\
\midrule
%% ---- N24 News ----
\multirow{4}{*}{\textbf{N24 News}}
& \%ACC $\uparrow$    & 98.5 & 97.3 & {\ul 98.1} & \textbf{98.3} & 97.7 & 98.5 & 97.9 & {\ul 98.3} & \textbf{98.5} & 98.2 \\
& c-RMSE $\downarrow$ & -- & \textbf{0.045} & 1.806 & 0.296 & {\ul 0.079} & -- & \textbf{0.044} & 1.723 & 0.125 & {\ul 0.057} \\
& CTL $\downarrow$    & -- & {\ul 0.027} & 0.388 & 0.377 & \textbf{0.005} & -- & {\ul 0.025} & 0.271 & 0.281 & \textbf{0.004} \\
& ICL $\downarrow$    & -- & \textbf{0.004} & 0.130 & 0.136 & {\ul 0.005} & -- & {\ul 0.025} & 0.061 & 0.085 & \textbf{0.003} \\
\midrule
%% ---- CUB200 ----
\multirow{4}{*}{\textbf{CUB200}}
& \%ACC $\uparrow$    & 91.3 & 66.2 & 69.8 & {\ul 70.2} & \textbf{79.3} & 95.8 & 72.3 & 76.7 & {\ul 77.8} & \textbf{85.3} \\
& c-RMSE $\downarrow$ & -- & \textbf{0.121} & 1.404 & 0.221 & {\ul 0.200} & -- & \textbf{0.081} & 1.528 & {\ul 0.201} & 0.273 \\
& CTL $\downarrow$    & -- & 0.032 & 0.063 & {\ul 0.029} & \textbf{0.026} & -- & \textbf{0.025} & 0.086 & 0.065 & {\ul 0.045} \\
& ICL $\downarrow$    & -- & 0.012 & 0.025 & {\ul 0.007} & \textbf{0.006} & -- & \textbf{0.006} & 0.026 & 0.012 & {\ul 0.010} \\
\midrule
%% ---- Avg. Rank ----
\multirow{4}{*}{\textbf{Avg. Rank}}
& \%ACC $\uparrow$    & -- & 4.00 & {\ul 2.25} & \textbf{1.25} & {\ul 2.25} & -- & 4.00 & \textbf{1.75} & \textbf{1.75} & {\ul 2.50} \\
& c-RMSE $\downarrow$ & -- & \textbf{1.00} & 4.00 & 3.00 & {\ul 2.00} & -- & \textbf{1.00} & 4.00 & {\ul 2.50} & {\ul 2.50} \\
& CTL $\downarrow$    & -- & {\ul 2.25} & 3.50 & 3.25 & \textbf{1.00} & -- & {\ul 2.25} & 3.25 & 3.25 & \textbf{1.25} \\
& ICL $\downarrow$    & -- & {\ul 1.75} & 3.25 & 3.50 & \textbf{1.50} & -- & {\ul 1.75} & 3.50 & 3.50 & \textbf{1.25} \\
\midrule
%% ---- Overall Avg. Rank ----
\multicolumn{2}{c|}{\textbf{Overall Avg. Rank $\downarrow$}}
& -- & {\ul 2.25} & 3.25 & 2.75 & \textbf{1.69} & -- & {\ul 2.25} & 3.12 & 2.75 & \textbf{1.88} \\
\bottomrule
\end{tabular}
\end{table*}
%%%%%%%%%%%%%%%%%%%% END TABLE 1 %%%%%%%%%%%%%%%%%%%%%

\subsection{Experimental Results}
\label{result}

\paragraph{Global Results.} Table~\ref{tab:results} shows the experimental results obtained from \method\ and its competitors on \texttt{CLIP-base} and  \texttt{CLIP-large} across the same training sets. \method\ approximately replicates the task accuracy of the black-box model and \texttt{CT-CBM} in most cases. \texttt{CT-CBM} and \texttt{Label-free} are consistently outperformed by \method\ and \texttt{Indep.-CBM} in concept detection quality (c-RMSE) and leakage reduction. \texttt{Indep.-CBM} achieves strong concept detection yet at the cost of degraded task accuracy. Moreover, \method\ surpasses all competing methods by a large margin on leakage score across all datasets and backbones, while \texttt{Label-Free} exhibits the largest concept detection errors. These results are further supported by the Pareto-optimal zone in Figure~\ref{fig:pareto_cbm}, where \method\ concentrates in the low-leakage, low-detection-error region. Taken together, \method\ achieves the best overall average rank across task accuracy, concept detection, and leakage metrics, under both CLIP-Base and CLIP-Large backbones (1.69 and 1.88, respectively), establishing it as the best trade-off between task accuracy and faithfulness overall.

%%%%%%%%%%%%%%%%%%%% TALBE 2 : PRINCIPAL RESULT %%%%%%%%%%%%%%%%%%%%%
\begin{table}[t]
\caption{Impact of \method\ enhancement of several CBM baselines with CLIP-base as a backbone.}
\label{tab:enhanced_baselines}
\centering
\footnotesize
\setlength{\tabcolsep}{4pt}
\begin{tabular}{@{}lcccccc@{}}
\toprule
& \multicolumn{3}{c}{\textbf{N24 News}} & \multicolumn{3}{c}{\textbf{DBpedia}} \\
\cmidrule(lr){2-4} \cmidrule(lr){5-7}
\textbf{Baseline} & \textbf{$\Delta$ Task} & \textbf{$\Delta$ Concept} & \textbf{$\Delta$ Task} & \textbf{$\Delta$ Task} & \textbf{$\Delta$ Concept} & \textbf{$\Delta$ Task} \\
& \textbf{Accuracy $\uparrow$} & \textbf{RMSE $\downarrow$} & \textbf{Leakage $\downarrow$} & \textbf{Accuracy $\uparrow$} & \textbf{RMSE $\downarrow$} & \textbf{Leakage $\downarrow$} \\
\midrule
\texttt{CT-CBM}     & $-0.3\%$ & $-50.4\%$ & $-69.9\%$ & $+8.7\%$ & $-38.3\%$ & $-96.6\%$ \\
\texttt{Indep.-CBM} & $-2.0\%$ & $+15.7\%$ & $-73.8\%$ & $-2.7\%$ & $-10.5\%$ & $-46.1\%$ \\
\bottomrule
\end{tabular}
\label{tab:baseline_enhancement}
\end{table}

%%%%%%%%%%%%%%%%%%%% END TABLE 2 %%%%%%%%%%%%%%%%%%%%%

\begin{figure}[t]
    \centering
    \includegraphics[width=0.999\textwidth]{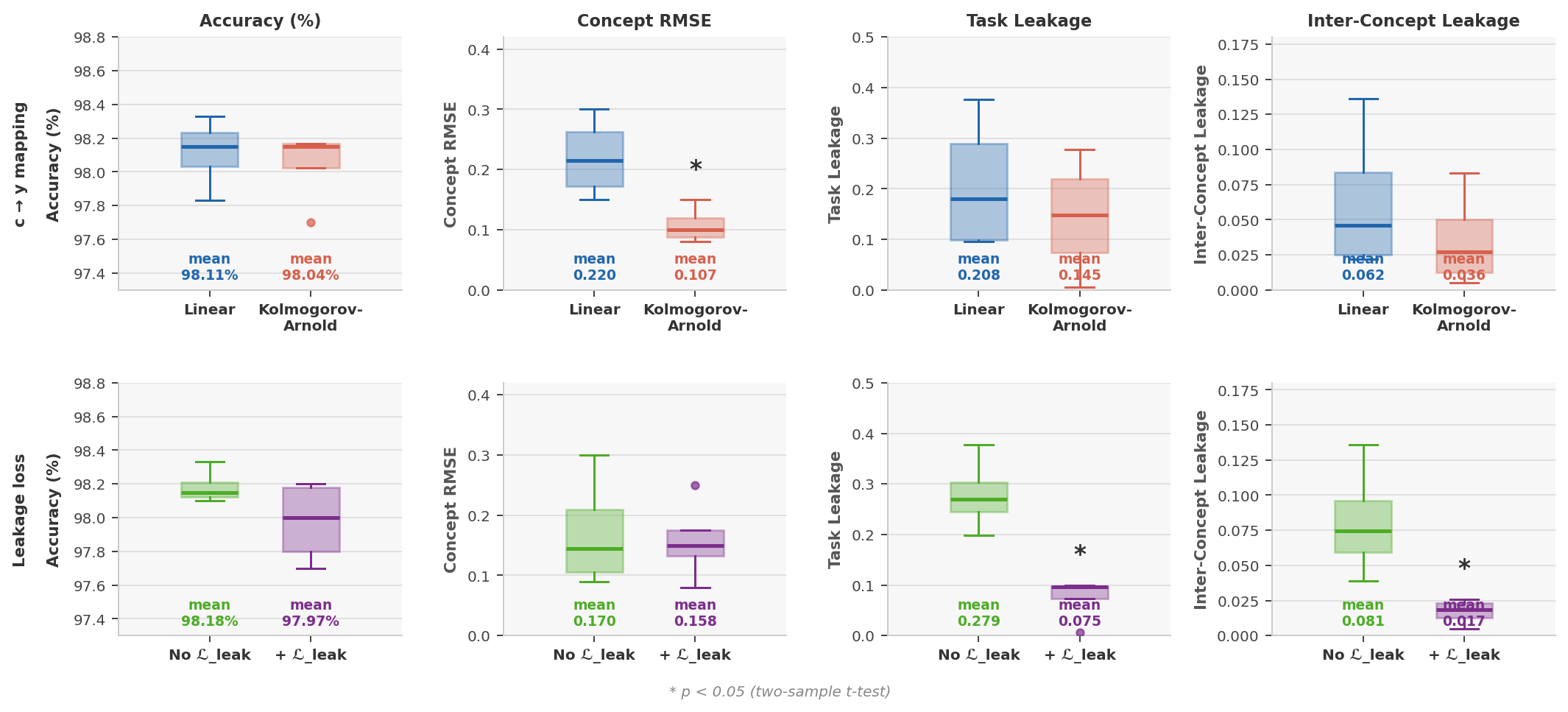}
    \caption{Ablation study of \method: effect of the KAN layer and the 
leakage loss on task accuracy, concept RMSE, and leakage (N24 dataset, CLIP-base backbone).}
    \label{fig:ablation_all}
\end{figure}

\paragraph{Baseline Enhancement.}

% BLABLBA commentaire baseline enhancement 
% Table~\ref{tab:baseline_enhancement}.
% LE TABLEAU ~\ref{tab:baseline_enhancement} montre que notre méthode peut s'emboiter avec les methodes existante  en améliorant leur capacité de detection et de reductio nde leakage dans presque tous les cas (attention baisse de detectio nde concept avec n24news de 15 pourcent) mais la perte est rien ocmparer au gain de faithfullness. 

% le gain sur la detection de concepts est mitigé mais la reductio ndu leakage est sûr avec des cas extrement important meme sur ind cbm qui avaient déjà une bonne detectio nde concept.
% On dégrae quasiment pas ou presque pas la performance, o nl'améliroe même sur dbpedia quand on ajoute notre methods on top of CT CBM tout en réduisant le leakage de 96 pourcent et améliorant la detectio nde concept de 50 POURCNET.
% On top of independant CBM? 

% generality : changer peut être ce terme

\begin{figure}[t]
    \centering
    \includegraphics[width=0.47\textwidth]{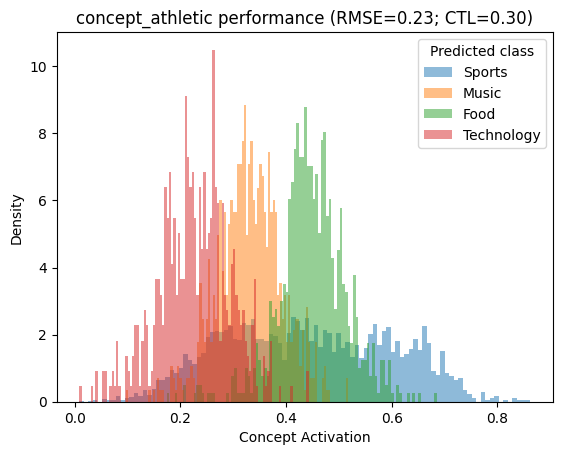}
    \includegraphics[width=0.47\textwidth]{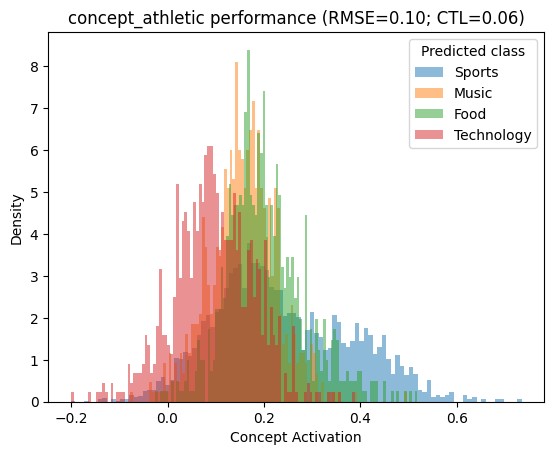}
    \caption{Effect of the leakage loss on concept activation distributions. Early in training (left), activations separate by predicted class, revealing concept-task leakage. Later (right), concept detection improved while the leakage loss reduces class information encoded in the activations, mitigating leakage.}
    \label{fig:leakage_distributions}
\end{figure}

% \begin{figure}[t]
%     \centering
%     \includegraphics[width=0.85\textwidth]{figures/kan_curves.png}
%     \caption{Illustration of the response curves produced by the KAN layer (Eq.~\ref{eq:response_curve}) for two classes of the N24News dataset (Sports and Music) for a random selection of concepts. The non-linear relationships learned by the KAN layer are clearly visible, reflecting the varying importance of each concept in predicting its corresponding final class.}
%     \label{fig:KAN_response_curves}
% \end{figure}

To further assess the generality of \method\ results, we evaluate the effect of integrating its components (leakage loss and KAN layer) into existing CBM competitors on N24News and DBpedia with CLIP-base as a backbone. As shown in Table~\ref{tab:baseline_enhancement}, incorporating \method’s components into \texttt{CT-CBM} and \texttt{Indep.-CBM} consistently and substantially reduces leakage, while resulting only in a relatively small decrease in task accuracy. When applied on top of \texttt{CT-CBM}, our approach preserves task accuracy on N24News and even improves it on DBpedia by $+8.7\%$ , while simultaneously improving concept detection up to $50\%$ on N24News and reducing leakage up to $-97\%$ on DBpedia. On top of \texttt{Indep.-CBM}, leakage is again strongly reduced ($-74\%$ and $-46\%$), at a modest task accuracy cost around $-2\%$. However, the effect on concept detection remains mixed: improved on N24News but marginally degraded on DBpedia, suggesting a tension between leakage reduction and concept detection in this setting. Overall, these results confirm that \method's components can be integrated into other mCBM pipelines to significantly improve faithfulness at a limited cost on task accuracy.

\paragraph{Ablation Study.}
We conduct an ablation study of \method\ examining two design choices: the concept-to-target mapping (linear vs.\ KAN) and the use of leakage regularization loss $\mathcal{L}_{\text{leak}}$. As shown in Figure~\ref{fig:ablation_all}, replacing the linear mapping with a KAN leaves task accuracy nearly unchanged ($98.11\%$ vs.\ $98.04\%$), while statistically significantly improving concept detection ($-51\%$ c-RMSE, from $0.220$ to $0.107$) and reducing both task leakage ($-30\%$, from $0.208$ to $0.145$) and inter-concept leakage. This suggests that KAN's non-linear expressiveness better disentangles concept predictions from the final task signal, alleviating the pressure on the concept layer to encode extraneous information. Adding $\mathcal{L}_{\text{leak}}$ to the training loss yields a statistically significant reduction in task leakage ($-73\%$, from $0.279$ to $0.075$), with negligible impact on concept detection (from $0.170$ to $0.158$) and a marginal accuracy cost (from $98.18\%$ to $97.97\%$). Crucially, although $\mathcal{L}_{\text{leak}}$ explicitly targets only task leakage, inter-concept leakage is also substantially reduced, empirically confirming the hypothesis formulated in Section~\ref{sec:explanatory_anaysis}: addressing task leakage alone can be sufficient to mitigate inter-concept leakage as well. Taken together, both components contribute complementary improvements in faithfulness at a negligible cost on task accuracy, validating our methodological choices.

% We conduct an ablation study of \method\ examining two design choices: the concept-to-target mapping (linear vs. KAN) and the use of leakage regularization loss $\mathcal{L}_{\text{leak}}$. As shown in Figure~\ref{fig:ablation_all}, replacing the linear mapping with a KAN leaves task accuracy nearly unchanged ($98.11\%$ vs. $98.04\%$), while statistically significantly improving concept detection ($-51\%$ c-RMSE, from $0.220$ to $0.107$) and reducing task leakage ($-30\%$, from $0.208$ to $0.145$) and inter-concept leakage. This suggests that KAN's non-linear expressiveness better disentangles concept predictions from the final task signal. Adding $\mathcal{L}_{\text{leak}}$ to the training loss yields a statistically significant reduction in task leakage (by $-73\%$, from $0.279$ to $0.075$), with negligible impact on concept detection (from $0.170$ to $0.158$) and a marginal accuracy cost (from $98.18\%$ to $97.97\%$). These results confirm the effectiveness of \method\ leakage regularization, and confirm that just adressing task leakage in the loss also enables to lower inter-concept leakage. Taken together, both components contribute complementary improvements in faithfulness at a negligible cost on task accuracy, validating the findings of Section~\ref{sec:explanatory_anaysis} and our methodological choices.

\paragraph{Addressing leakage}

To illustrate how leakage can render a model unfaithful with respect to its concept predictions, we show in Figure~\ref{fig:leakage_distributions} the concept activation distributions for the concept \textit{athletic performance} at early training epochs, when concept error is still large and the concept task leakage loss has not yet taken effect. One can clearly observe that the concept is more strongly activated for the class Sports, as expected; however, for classes in which this concept should contribute minimally (Technology, Music, and Food), the activation distributions are clearly separated from one another, despite having no justification for \textit{athletic performance} to be more activated in Food-related articles than in others. At a later stage of training (right-hand side of Figure~\ref{fig:leakage_distributions}), concept error and leakage decreased, and the distributions for these three classes have nearly merged, indicating that the concept activation is much less class-discriminative, which is the expected faithful behavior.

\paragraph{KAN Layer Interpretation}

\begin{figure}[t]
    \begin{minipage}[t]{0.35\linewidth}
        \centering
        \includegraphics[width=\linewidth]{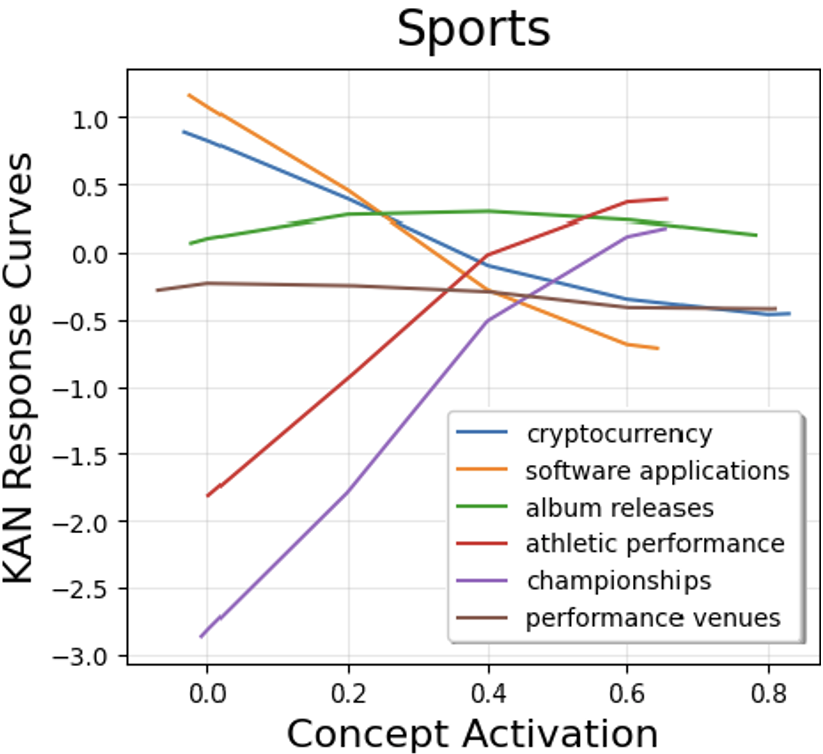}
        \caption{KAN response curves for Sports class (N24News).}
        \label{fig:KAN_response_curves}
    \end{minipage}
    \hfill
    \begin{minipage}[t]{0.63\linewidth}
        \centering
        \includegraphics[width=\linewidth]{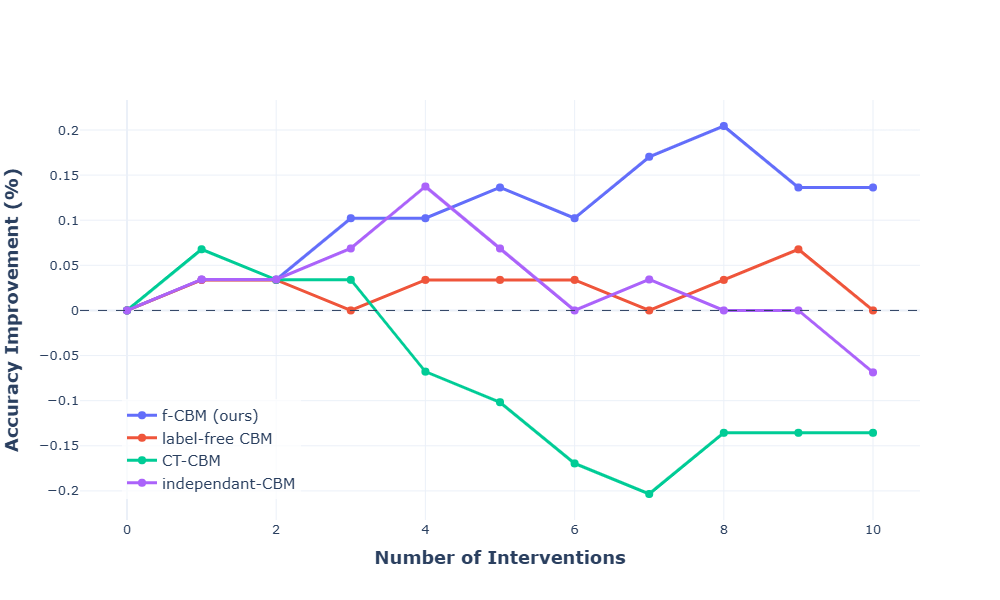}
        \caption{Intervention acc. improvements for different CBM configurations on N24News with CLIP-base.}
        \label{fig:intervention}
    \end{minipage}
\end{figure}

As studied in \cite{parisini2025leakage}, a misspecified final head can 
cause leakage in CBMs, as the model encodes additional information into concept activations to compensate for a limited classification capacity. To illustrate this non-linear behavior, we show in Figure~\ref{fig:KAN_response_curves} the response curves from Eq.~\ref{eq:response_curve}, which describe how each concept activation translates into a contribution toward each final class. For instance, for the class Sports, concepts such as \textit{athletic performance} and \textit{championships} exhibit positive response curves, meaning higher concept activation increasingly contributes to predicting that class, while concepts associated with other classes display flat or negative response curves. The non-linear behavior is further evidenced by the saturation of response curves at large concept activation values, indicating that beyond a certain threshold, further increases in concept activation no longer strengthen the corresponding class prediction.

\paragraph{Intervention Analysis.}

% \begin{figure}[t]
%     \centering
%     \includegraphics[width=0.7\linewidth]{figures/intervention_plot.png}
%     \caption{Intervention acc. improvements for different CBM configurations on  N24 on CLIP-base.}
%     \label{fig:intervention}
% \end{figure}

Beyond standard metrics, concept intervention provides an operational test of faithfulness: if a CBM truly relies on concept semantics, replacing predicted activations with ground-truth values should improve predictions. Conversely, degradation under intervention reveals that the model exploits leaked information destroyed by the correction. We follow the protocol of~\cite{cbm_intro}, ranking concepts by validation-set accuracy gain and intervening sequentially on the test set. As shown in Figure~\ref{fig:intervention}, \texttt{f-CBM} is the only method that consistently benefits from expert corrections, achieving accuracy gains as more concepts are intervened on. This is a direct consequence of its low leakage: because concept activations encode only their intended semantics, substituting ground-truth values is a semantically coherent operation that strictly refines the input to the prediction head. In contrast, \texttt{CT-CBM} accuracy degrades under intervention, revealing that its CBL encodes task-relevant signals beyond concept semantics that are disrupted when activations are overwritten. \texttt{Label-free} CBM shows near-zero improvement, consistent with its high concept detection error. \texttt{Independent-CBM} initially follows a trend similar to \texttt{f-CBM} but plateaus and slightly degrades after several interventions. These results confirm the findings of~\cite{parisini2025leakage} linking leakage to intervention performance, showing the importance of properly addressing faithfulness in CBMs overall.

\section{Conclusion}
\label{conclusion}

We introduced \method, a multimodal Concept Bottleneck Model framework that jointly addresses concept detection quality and information leakage, two core components of faithfulness in CBMs. By leveraging CLIP as a unified vision-language backbone, \method\ handles both image and text inputs within a single concept bottleneck architecture. Our two complementary strategies, a differentiable leakage-aware training objective and a KAN-based prediction head, target faithfulness from different angles: the former explicitly penalizes task-relevant information encoded beyond concept semantics, while the latter provides sufficient expressiveness to reduce the pressure on the concept layer to smuggle additional signals. Experiments show that \method\ achieves the best overall trade-off between task accuracy and faithfulness, consistently obtaining the lowest leakage scores among all competitors. Crucially, \method\ is the only method that reliably benefits from concept interventions, confirming that faithfulness is not only a desirable theoretical property but a practical prerequisite for the practical use of CBMs by humans. Beyond global predictions, the learned KAN response curves offer an interpretable, per-concept view of the decision process, revealing non-linear and saturating relationships between concept activations and class predictions..

\bibliographystyle{splncs04}
\bibliography{compit}

% no appendix for ECML
%\appendix
%\section{Appendix}
%\label{sec:appendix}
%\subsection{Scientific Libraries}
%We used several open-source libraries in this work: pytorch~\cite{paszke2019pytorch}, HuggingFace transformers~\cite{wolf2020transformers} sklearn~\cite{pedregosa2011scikit} and Captum~\cite{miglani_using_2023}. 

%\subsection{Autoregressive language models implementation details}
%\label{sec:appendix_slm_implementation_details}
%\paragraph{Language Models.} The library used to import the pretrained autoregressive language models is Hugging-Face. In particular, the backbone version of Gemma-2-9B is \texttt{gemma-2-9B-it}.

\end{document}